\documentclass{article}

\usepackage{arxiv}

\usepackage[utf8]{inputenc} 
\usepackage[T1]{fontenc}    
\usepackage{hyperref}       
\usepackage{url}            
\usepackage{booktabs}       
\usepackage{amsfonts}       
\usepackage{nicefrac}       
\usepackage{microtype}      
\usepackage{lipsum}
\usepackage{graphicx}
\usepackage{color}
\usepackage{tablefootnote}
\graphicspath{ {./images/} }

\title{An overview of event extraction and its applications}

\author{
 Jiangwei Liu \\
  School of Information Management and Engineering\\
  Shanghai University of Finance and Economics\\
  Shanghai 200433, China \\
  \texttt{majorliujw@gmail.com} \\
   \And
 Liangyu Min \\
  School of Information Management and Engineering\\
  Shanghai University of Finance and Economics\\
  Shanghai 200433, China \\
  \texttt{minux@163.sufe.edu.cn} \\
  \And
 Xiaohong Huang \\
  School of Information Management and Engineering\\
  Shanghai University of Finance and Economics\\
  Shanghai 200433, China \\
  \texttt{huangxiaohong@163.sufe.edu.cn} \\

}

\begin{document}
\maketitle
\begin{abstract}
With the rapid development of information technology, online platforms have produced enormous text resources. As a particular form of Information Extraction (IE), Event Extraction (EE) has gained increasing popularity due to its ability to automatically extract events from human language. However, there are limited literature surveys on event extraction. Existing review works either spend much effort describing the details of various approaches or focus on a particular field. This study provides a comprehensive overview of the state-of-the-art event extraction methods and their applications from text, including closed-domain and open-domain event extraction. A trait of this survey is that it provides an overview in moderate complexity, avoiding involving too many details of particular approaches. This study focuses on discussing the common characters, application fields, advantages, and disadvantages of representative works, ignoring the specificities of individual approaches. Finally, we summarize the common issues, current solutions, and future research directions. We hope this work could help researchers and practitioners obtain a quick overview of recent event extraction.
\end{abstract}


\section{Introduction}
With the rapid development of information technology, electronic textual data generated by the Internet provide a resource of unbounded information-bearing potential. Over the years, Information Extraction (IE) has gained increasing popularity because it helps exploit this potential by automatically extracting content from human language \cite{DoddingtonMitchell-1}. Event Extraction (EE) originated in the late 1980s when the U.S. Defense Advanced Research Projects Agency (DARPA) boosted research into message understanding \cite{HogenboomFrasincar-2}. Now event extraction has become an important and challenging task, which aims to discover event triggers with specific types and their arguments \cite{ChenXu-3}. 

Event extraction plays an important role in many applications in various fields. In the security field, Tanev et al. \cite{TanevPiskorski-4} perform real-time news event extraction for global crisis monitoring. In the intelligent transportation field, Sakaki et al. \cite{SakakiMatsuo-5} develop a system that extracts real-time driving information using social media to provide important events for drivers, such as traffic jams and weather reports. Sheng et al. \cite{ShengGuo-17} study the overlapping event extraction problem in the financial field, and Zheng et al. \cite{ZhengCao-6} propose a novel end-to-end document-level event extraction framework from Chinese financial announcements. In the social media field, Ritter et al. \cite{RitterEtzioni-7}, Zhou et al. \cite{ZhouChen-8}, Kunneman and Van Den Bosch \cite{KunnemanVan-Den-Bosch-9}, and Peng et al. \cite{PengLi-10} develop novel open-domain event extraction models to extract events from Twitter. In the biomedical field, there is much research that extracts medications and associated adverse drug events (ADEs) from clinical documents \cite{WeiJi-11, LiuZhao-12, PengMoh-13}. In the legal field, extracting events from court decisions can provide a visual overview of what happened throughout a case by representing the main legal events, together with relevant temporal information \cite{FiltzNavas-Loro-14}. Many studies have focused on proposing new approaches to tackle the challenges of general event extraction \cite{BuykoFaessler-15, HennSticha-16, ShengGuo-17,WenLin-18, VeysehNguyen-19}.

\begin{figure}[htbp]
	\centering
	\includegraphics[width= 6.5 in]{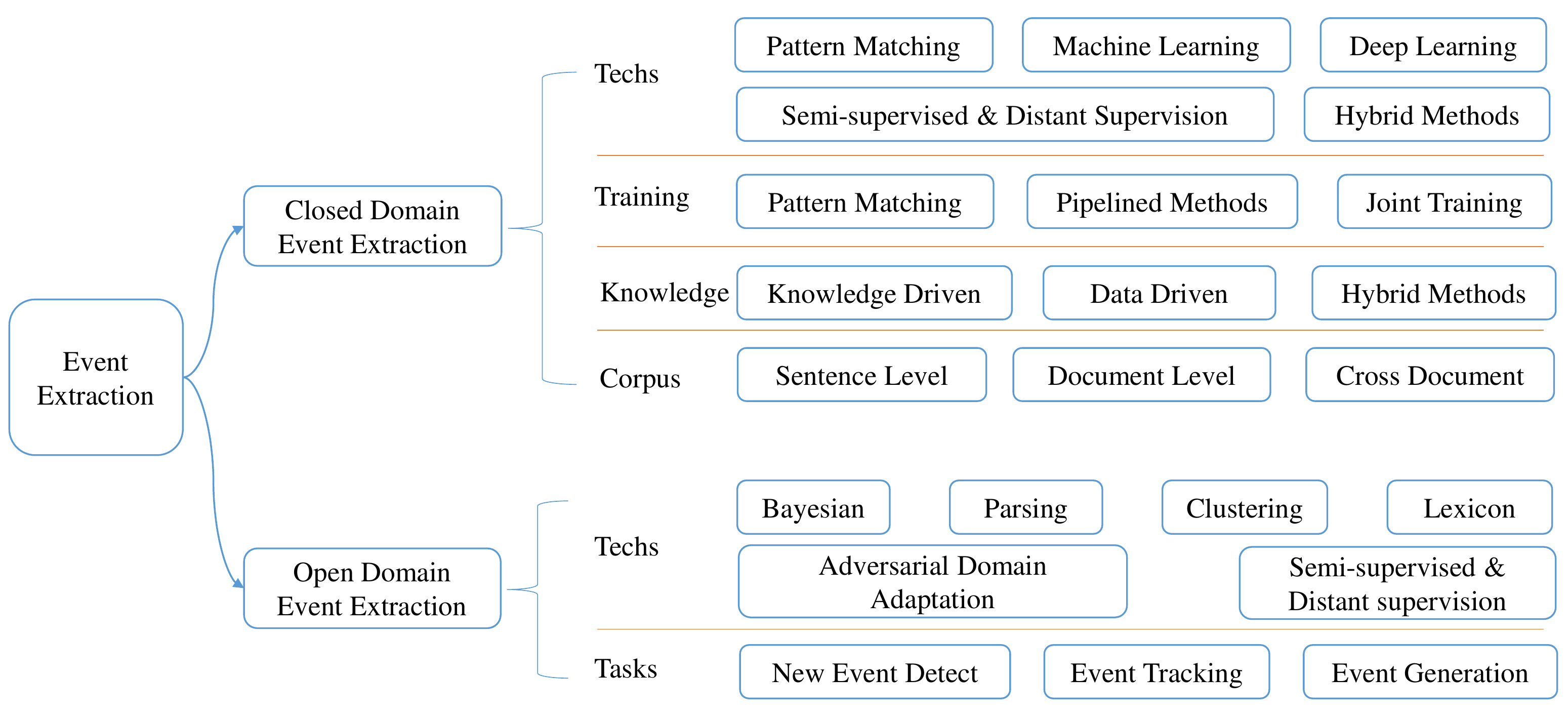}	
	\caption{Different Event Extraction classification.}
	\label{fig1}	
\end{figure}

According to different classified methods, the existing event extraction literature can be classified into different categories. We summarize the typical research works and categorize them in Figure \ref{fig1}. Event extraction includes closed-domain and open-domain event extraction, which are two mainstreaming parts. The former aims to discover event triggers with specific types and their arguments, whereas the latter concentrates upon detecting new events or tracking the change of state of a known event. This study mainly summarizes the literature from the technique view, with other classification methods as supplementary.

\paragraph{Closed-domain event extraction.} 
From the view of techniques used, existing approaches can be divided into four categories: pattern matching, machine learning, deep learning, and semi-supervised learning methods. It is worth noting that semi-supervised learning methods are separately treated as a single category because much research has recently used semi-supervised or distant learning methods to enhance corpora and has become a research hot. 

From the view of how to train a model, existing approaches can be categorized into pattern matching, pipelined training, and joint training methods. Which manner is chosen mainly depends on how to treat the subtasks of event extraction by researchers.

From the view of whether much expert knowledge is needed, existing approaches can be divided into knowledge-driven, data-driven, and hybrid methods \cite{HogenboomFrasincar-20}. Knowledge-driven methods usually need expert knowledge to design delicate patterns. Data-driven approaches mainly exploit knowledge from big data through statistics or deep learning methods. The hybrid approaches combine the above mentioned methods. 

Existing research can be divided into sentence level, document level, and cross-document level from the corpus level on which the event extraction tasks are performed.

\paragraph{Open-domain event extraction.}
Open-domain event extraction is highly different from closed-domain event extraction because it focuses on detecting new or unexpected events from texts. So there are no predefined event types, and event schema induction is a critical subtask of open-domain event extraction. From the view of technologies used, existing approaches can be divided into Bayesian-based \cite{WangZhou-21}, clustering-based \cite{PengLi-10}, parsing-based \cite{RitterEtzioni-7}, lexicon-based \cite{de-VroeGuillou-22}, semi-supervised \cite{VeysehNguyen-19} and distant supervision based \cite{FiltzNavas-Loro-14}, Adversarial Domain Adaptation based \cite{NaikRose-23}. From the view of the task target, the existing research can be categorized into new event detection, event generation, and event tracking.

Despite the importance and popularity of event extraction, there are limited comprehensive reviews and summaries on the recent study of event extraction \cite{HogenboomFrasincar-20, ZhanJiang-24, XiangWang-25}. Most of the surveys research mainly focus on some specific field, for example, deep learning schema-based event extraction \cite{LiPeng-26}, multilingual event extraction \cite{DanilovaAlexandrov-27}, event extraction from social networks \cite{MejriAkaichi-28}, biomolecular event extraction \cite{VanegasMatos-29, Shahab-30}, event extraction for decision support systems \cite{HogenboomFrasincar-2}, etc. Another limitation is that most existing surveys, including comprehensive reviews, lack a summary of recent open-domain event extraction research. From this view, we review and provide an overview of recent event extraction literature. Different from previous survey research, we summarize the contributions of this study as follows: 

(1) We systematically review the literature of event extraction from the technique view, both closed-domain and open-domain event extraction included. In each section, we review the models, techniques, event levels, datasets, and application fields of the representative research and summarize them in a corresponding table by year. 

(2) A trait of this survey is that we try to provide an overview in moderate complexity. We ignore the specificities of individual research and avoid discussing the details of the individual research. We focus on discussing the common characters, application fields, advantages, and disadvantages of representative works. We hope this work could help researchers and practitioners obtain a quick outline of recent event extraction. 

(3) We summarize the common issues and challenges that hinder event extraction generalization and industrial applications. And currently corresponding solutions and research directions are also mentioned in the following.

The remainder of this paper is organized as follow. We first introduce event extraction task definition, commonly used corpora, and evaluation metrics. Then review and summarize the literature in the technique view, with closed-domain event extraction in section 3 and open-domain event extraction in section 4. Section 5 summarizes and discusses the current common research issues and future directions. Conclusions are followed in section 6.

\section{Event Extraction}
\subsection{Event Extraction Task Definition}
\begin{figure}[htbp]
	\centering
	\includegraphics[width= 6.5 in]{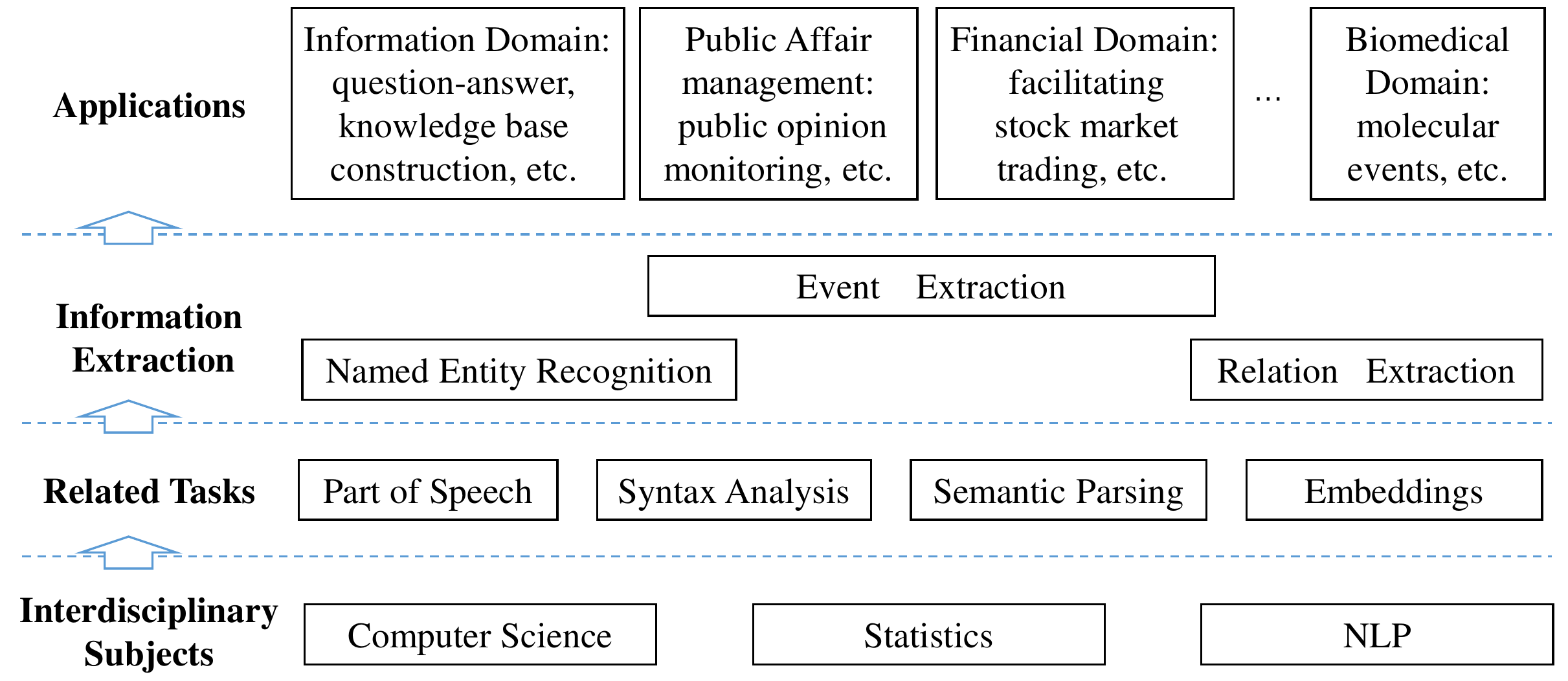}	
	\caption{Demonstration of the relationship between event extraction and other interdisciplinary subjects and techniques.}
	\label{fig2}	
\end{figure}

As a particular form of information, event extraction involves named entity recognition (NER) and relation extraction (RE), and mostly depends on the results of these tasks. As an interdisciplinary subject, event extraction is closely related to computer science, statistics, and natural language processing. We demonstrate the relations from its fundamentals to its applications in Figure \ref{fig2}.

Following the event extraction task definition in ACE 2005, an event is frequently  described as a change of state, indicating a specific occurrence of something that happens in a particular time and a specific place involving one or more participants. It can help answer the "5W1H" questions, i.e., "who", "when", "where", "what", "why" and "how" about an event. ACE employs the following terminologies to describe an event extraction task:
\paragraph{Event mention}: An Event mention usually is a phrase or sentence that describes an event in which a trigger and corresponding arguments are included.
\paragraph{Event trigger}: It usually is a verb or a noun that most clearly expresses the core meaning of an event.
\paragraph{Event type}: It refers to the category to which the event corresponds. In most cases, event types are predefined manually, categorized by event triggers. For instance, there are eight event types and 33 subtypes predefined in the ACE 2005 event corpus. While in open-domain event extraction, it is not predefined explicitly but usually can be represented by the event trigger.
\paragraph{Event argument}: Event arguments are the main attributes of events. They are usually entity mentions describing the event state change, involving who, what, when, where, and how.
\paragraph{Argument role}: An argument role is a function or position that an event argument performs in the relationship between the event argument and the trigger.

\begin{figure}[htbp]
	\centering
	\includegraphics[width= 6.5 in]{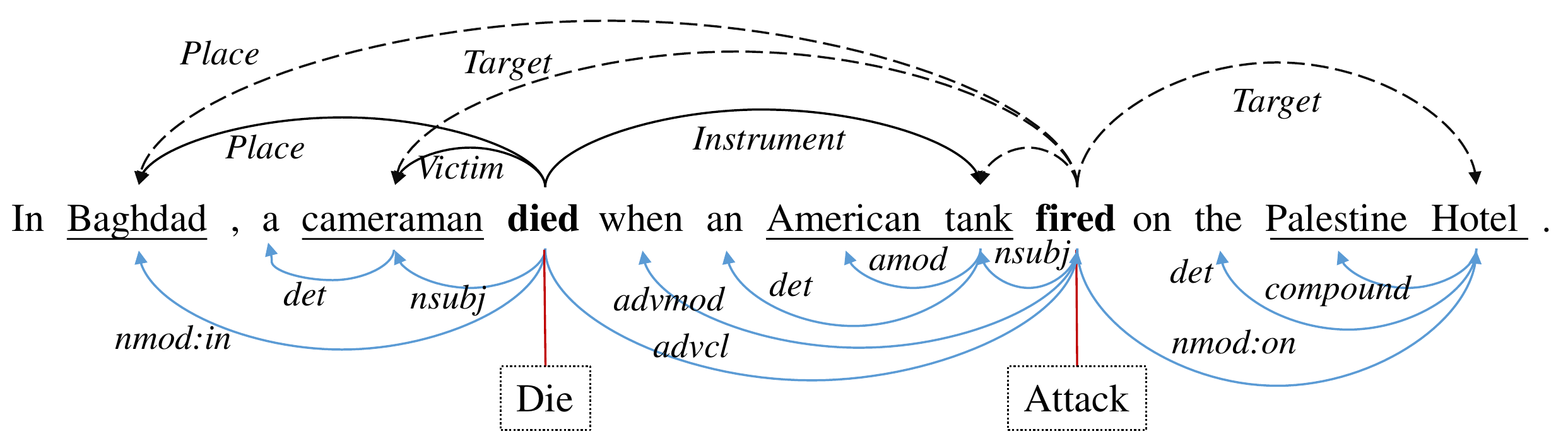}	
	\caption{An example of two events in one sentence: Die and Attack. The upper arcs link event triggers to their corresponding arguments, with the argument roles on the arcs. The lower side demonstrates the syntactic parser results.}
	\label{fig3}	
\end{figure}

For example, there are two event types involved in sentence S1: "Die" and "Attack", triggered by "died" and "fired", respectively. For Die event, "Baghdad", "cameraman", and "American tank" are its arguments with corresponding roles: Place, Victim, and Instrument, respectively. For Attach event, "Baghdad", "cameraman", "American tank" and "Palestine Hotel" are its arguments with corresponding role: Place, Victim, Instrument and Target, respectively. This is a somewhat more complex example with three arguments shared, which is more challenging than the simple case with one event type in one sentence. Figure \ref{fig3} shows the event extraction annotation and the syntactic parser results.
\begin{itemize}
	\item  \emph{S1: In Baghdad, a cameraman died when an American tank fired on the Palestine Hotel.}
\end{itemize}

The closed-domain event extraction task can be divided into four subtasks: trigger identification, event type classification, argument identification, and argument role classification. From the manner of how to organize the subtasks of the event extraction, most of the existing closed-domain event extraction methods can be divided into two mainstreaming categories: pipelined-based method and joint-based method. The pipeline-based method utilizes the idea of Divide-and-Conquer algorithms; thus, the advantage is that it simplifies each subtask and can afford information for subsequent subtasks. In contrast, the disadvantages are that it propagates cascading errors, and the overall performance dramatically relies on the previous subtasks. The joint-based method considers the subtasks independently, thus does not propagate errors among the subtasks. Accordingly, the disadvantages are that it can not utilize previous subtasks' information and needs more large-scale delicate labeled data to train the models.

\subsection{Event Extraction Corpora}
Event Extraction corpora are annotated by professionals or experts with domain knowledge and used to train or evaluate models. This section mainly introduces some representative event extraction corpora afforded by public evaluation programs or mentioned in previous literature. We summarize these popular corpora in Table \ref{tab1}.

\begin{table}
	\caption{Summary of existing representive Event Extraction Corpora.}
	\centering
	\begin{tabular} {llllll} 
		\toprule 
		Corpus	& Languages	& Event Types	& Documents/Sentences	& Events\\
		\midrule 
		ACE 2005	& English/Chinese/Arabic	& 8	& 599/633	& 6000\\
		TimeBANK	& English	& 8	& 183/300	& 7935\\
		Factbank	& English	& 12	& 208/-	& 9488\\
		GENIA	& English	& 47	& 1999/18545	& 36114\\
		DTD	& English/Chinese	& -	& 134000/-	& -\\
		GNBusiness	& English	& 8	& 680 (12985)/-	& -\\
		ASTRE	& English	& 12	& 100 (1038)/42600	& -\\
		Patch Hate Crimes	& English	& 3	& 11130	& -\\
		CEC	& Chinese	& 5	& 332/-	& 5991\\
		DuEE 2020	& Chinese	& 65	& -/17000	& 20000\\
		
		\bottomrule 
	\end{tabular}
	\label{tab1}
\end{table}

\begin{itemize}
\item The ACE 2005 event corpus contains eight event types and 33 subtypes, with about 6000 labeled examples in 599 documents (633 Chinese documents). Events in the ACE 2005 corpus are represented in terms of their attributes and their participants. The participants are the ACE entities that participate in the event. ACE events are, in essence, a generalization of ACE relations \cite{DoddingtonMitchell-1}.

\item The texts in the TimeBANK corpus \cite{PustejovskyHanks-31} cover various media sources of the news domain. It was a gold-standard human-annotated corpus marked up for temporal expressions, events, and temporal relations holding between events and events times, following the TimeML (Time Markup Language) annotation scheme. The TimeBank 1.2 \cite{PustejovskyLittman-32} contains 183 articles with 27592 TimeML tags, among which 7935 are event tags.

\item The Factbank corpus \cite{SauriPustejovsky-33} is built on TimeBank 1.2 and part of the AQUAINT TimeML corpus. The difference is that the Factbank corpus is supplemented with additional information concerning the factuality of events. It consists of 208 documents and contains a total of 9488 manually annotated events.

\item The GENIA corpus \cite{KimOhta-34} is a semantically annotated corpus of biological literature. The GENIA corpus 3.0 consists of 1999 abstracts taken from the MEDLINE database. The current GENIA corpus event annotation covers 1,000 of the 1999 abstracts of the primary GENIA corpus, marking 36114 events in 9372 sentences. The more detailed event annotation information can be found at http://www.geniaproject.org/genia-corpus/event-corpus.

\item The TDT corpora \footnote{\url{https://catalog.ldc.upenn.edu/byproject\#TDT-corpora}}  \cite{PapkaAllan-35, Allan-36} are used for Topic Detection and Tracking research programs, including TDT Pilot, TDT2, and TDT3 corpus. The TDT Pilot corpus contains approximately 16,000 stories and 25 events. The TDT2 corpus contains over 74,000 stories with more than 100 topics. The TDT3 corpus 2.0 contains over 31200 English stories and 12800 Chinese stories. They are usually used in open-domain event extraction tasks: to detect the occurrence of new events (detection) and track the reoccurrence of old events (tracking).

\item The GNBusiness dataset is a large-scale dataset annotated with diverse event types and explainable event schemas, released along with the ODEE (Open Domain Event Extraction) algorithm \cite{LiuHuang-37}. It contains 55618 business news reports with 13047 news clusters in 288 batches from Oct. 17, 2018, to Jan. 22, 2019, among which 680 clusters are annotated.

\item The ASTRE corpus \cite{NguyenTannier-107}, dedicated to the evaluation of event schema induction, contains 1038 documents. There are 100 documents selected from Wikinews inside the category Laws \& Justice. The rest documents are retrieved by the Google search engine, similar to the 100 initial seed documents. Only the Wikinews documents are manually annotated to evaluate model performance, while the others are left for unsupervised learning.

\item The Patch Hate Crimes corpus \cite{DavaniYeh-38} includes hyper-local news articles from 1217 cities based in the USA, scraped from the Patch \footnote{\url{https://www.patch.com}} website in the "Fire and Crime" category. A subset of 11130 articles are annotated with a binary label (whether the article represents a specific hate crime) and attributes (including eight targets and four types of hate crime actions).

\item The CEC corpus \footnote{\url{https://github.com/shijiebei2009/CEC-Corpus}} (Chinese Emergency corpus) collects breaking news events reported in Chinese, with a total of 332 documents. It consists of five event types: earthquake, fire, traffic accident, terrorist attack, and food poisoning.

\item The DuEE 2020 corpus  \footnote{\url{http://lic2020.cipsc.org.cn/}} is released by Baidu and adopted in Language and Intelligent Technology Competition 2020. The corpus is selected and determined according to the hot search board of Baidu. It consists of 17,000 sentences containing 20,000 events of 65 event types.
\end{itemize}

In summary, although there have been various annotated event extraction corpora, including closed-domain and open-domain corpora, there are still many limitations. Firstly, from the domain view, most of the existing corpora are developed for closed-domain tasks with limited event types. Secondly, from a corpus size view, most corpora are small and sparse because the annotation is such a cost-prohibitive process. Thirdly, from a utility view, the reusability of existing corpora greatly depends on the targeted domains. Lastly, there is still a lack of generally acknowledged large-scale corpus for open-domain event extraction.

\subsection{Event Extraction Evaluation Metrics}

The event extraction task, especially the closed-domain event extraction task, can be regarded as a classification task or Sequence Labeling task. Most existing literature uses classification metrics to evaluate the event extraction performance. In accordance with IE and TM, performance is generally measured by calculating the quantity of true positives and negatives, as well as that of false positives and negatives. The most used metrics, e.g., precision, recall, and F1 score, are calculated as follows:

\begin{equation} 
	Precision = \frac{{TP}}{{TP + FP}}
\end{equation}
\begin{equation} 
	Recall = \frac{{TP}}{{TP + FN}}
\end{equation}
\begin{equation} 
	F1 = \frac{{2*Precision*Recall}}{{Precision + Recall}} = \frac{{2*TP}}{{2*TP + FP + FN}}
\end{equation}

These performance measures provide a brief explanation of the "Confusion Metrics". True positives (TP) and true negatives (TN) are the observations that are correctly predicted. In contrast, false positives (FP) and false negatives (FN) are the values that the actual class contradicts with the predicted class.

Open-domain event extraction aims to detect the unreported events or track the progress of the previously spotted events. In most cases, it has no predefined schemas and event types. But with the help of annotated corpus, it still can be transferred into a classification problem and thus uses the mentioned evaluation metrics. Many works conduct the open-domain event extraction by clustering algorithms, and therefore, some clustering evaluation metrics like mutual information or Chi-Square are often employed. For example, normalized pointwise mutual information (nPMI) can be used to measure the slot coherence \cite{LiuHuang-37}:

\begin{equation} 
	nPMI(x,y) = \frac{{\log \frac{{f(x,y)}}{{f(x)*f(y)/W}}}}{{\log \frac{1}{{f(x,y)/W}}}}
\end{equation}

where $W$ is the total number of words in the corpus; $f(x)$ and $f(y)$ are frequencies of $x$ and $y$ in the corpus; $f(x,y)$ is the occurrence frequency of word pair $(x,y)$ in the corpus. There are also other variants, e.g., ${\rm{c}}PMI$ (Corpus Level Significant PMI) and $PM{I^2}$ used in the literature \cite{Damani-39, Pecina-40}.

\section{ Closed Domain Event Extraction}

This section categorizes closed-domain event extraction approaches into pattern matching, machine learning, deep learning, and semi-supervised learning methods. The categorical arrangement also considers and follows the time when the technique became a popular mainstream. We focus on providing an overview of closed-domain event extraction by concentrating on the most common characters, including the main idea, common framework, applicated area, advantage, and disadvantage. Many peculiarities of individual approaches are not considered in this study.

\subsection{Pattern Matching based Methods}
One character of pattern matching based methods is that they depend on domain-specific event templates, which require a great deal of manual knowledge engineering to construct elaborately designed features. The earliest event extraction methods were mainly based on syntax trees or regular expressions. 

\begin{figure}[htbp]
	\centering
	\includegraphics[width= 6.5 in]{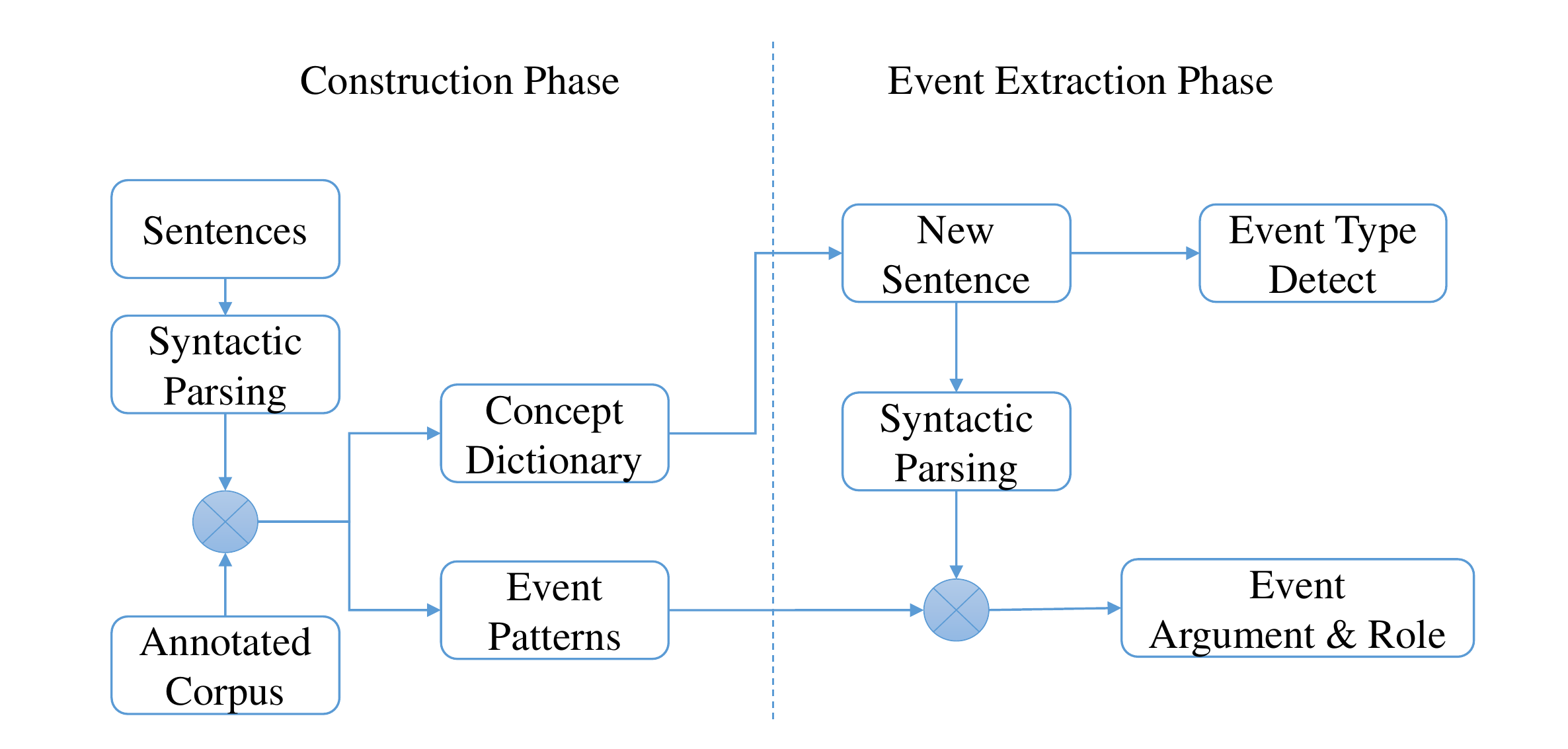}	
	\caption{Demonstration of a typical pattern matching based event extraction process.}
	\label{fig4}	
\end{figure}

The typical representative work might be the AutoSlog system, developed by Ellen in 1993 \cite{Riloff-41}. It first defines 13 linguistic patterns with the help of a conceptual sentence analyzer. These linguistic patterns are used to build a domain-specific dictionary of concepts automatically. Then the AutoSlog uses the trigger word dictionary to detect a potential event. Lastly, it associates the event patterns and linguistic features, e.g., part-of-speech tags (POS) generated by the sentence parser, to assemble the argument and its corresponding role. We summarize this typical process in Figure \ref{fig4}.

Due to its outstanding performance in specific domains, the research of pattern matching based event extraction has exploded in various fields, such as the biomedical \cite{YakushijiTateisi-42, KilicogluBergler-43, BuykoFaessler-15}, general information extraction \cite{YangarberGrishman-44, LeeChen-45}, finance and economics \cite{BorsjeHogenboom-46}, etc. Akane et al. \cite{YakushijiTateisi-42} design a program to extract events from biomedical papers using a full parser. Halil et al. \cite{KilicogluBergler-43} use syntactic dependency and rules to perform biological event extraction. Ekaterina et al. \cite{BuykoFaessler-15} incorporate manually curated dictionaries and machine learning methodologies to extract event triggers and arguments on trimmed dependency graph structures. Roman et al. \cite{YangarberGrishman-44} propose an automatic event pattern discovery approach, which can identify a set of relevant documents and a set of event patterns from un-annotated text, starting from a small set of "seed scenario patterns". Chang et al. \cite{LeeChen-45} propose a method that can effectively summarize the Chinese e-news by four main components: Chinese POS tagger, Chinese term filter, Event Ontology Filter, and Summarization Agent. Jethro et al. \cite{BorsjeHogenboom-46} propose the use of lexico-semantic patterns for financial event extraction from RSS news feeds.

The typical characters lie in two aspects: (1) utilizing lexical features, e,g., part-of-speech tags (POS), entity information, and morphology features (token, lemma, etc.); (2) utilizing delicate event patterns normally designed by experts with domain knowledge. 

Several advantages of pattern-based approaches are summarized as follow. First, it needs less corpus than data-driven methods. Second, it has better interpretability due to its patterns are manually designed and maintained. Third, it can achieve high extraction accuracy in a specific domain once the patterns are well designed.

We summarize the disadvantages of pattern-based approaches from designing and generalizing views. First, developing and maintaining the delicate event patterns is rather time-consuming and labor-intensive. Second, because pattern designing is strongly dependent on the expression form of text, it needs much effort to transfer the patterns from one domain to another. Low reusability of designed patterns or templates limits its generalization.

\subsection{Machine Learning based Methods}
\begin{figure}[htbp]
	\centering
	\includegraphics[width= 6.5 in]{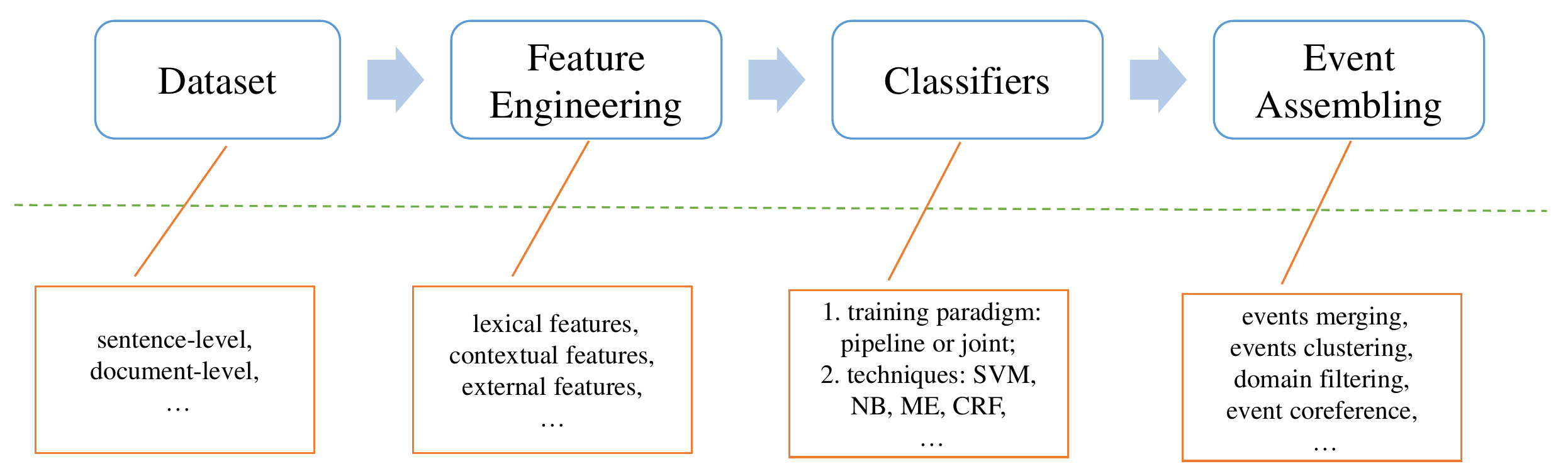}	
	\caption{Demonstration of an abstract process of machine learning based event extraction approaches. The lower side demonstrates the usually executed steps in the corresponding phases.}
	\label{fig5}	
\end{figure}

To alleviate the difficulty in designing delicate event patterns, many researchers have explored machine learning methods to extract events. In this section, we first review the typical machine learning based event extraction literature and summarize it in Table 2 from the view of the year, model, paradigm, technique, datasets used, event-level performed, and application area. We also summarize and plot the typical abstract process in Figure \ref{fig5}. Then we focus on discussing the common characters of typical research from feature engineering, the paradigm, technique, and application fields, without considering spending much effort in describing the details of specific methods. We finally summarize the advantages and disadvantages of machine learning based event extraction methods.

\begin{table}
	\caption{Summary of representative machine learning based event extraction methods.}
	\centering
	\scriptsize
	\begin{tabular} {llllllll} 
		\toprule 
		Year	& Model	& Paradigm	& Techniques	& Datasets 	& Event Level	& Application Field\\
		\midrule 
		2021	& Henn et al. \cite{HennSticha-16}		& pipeline	& SVMs	& ToMK	& document-level	& general IE\\
		2016	& Peng et al. \cite{PengMoh-13}		& pipeline	& SVM, Naive Bayes	& Dataset \cite{PengMoh-13}		& document-level	& biomedical\\
		2012	& Lu and Roth \cite{LuRoth-60}		& joint	& ME, Markov CRF	& ACE2005	& sentence-level	& general IE\\
		2012	& Chen and Ng \cite{ChenNg-59}		& joint	& SVM	& ACE2005	& sentence-level	& general IE\\
		2012	& Li et al. \cite{LiZhu-58}		& joint	& ILP, ME, CRF	& ACE2005	& sentence-level	& general IE\\
		2012	& Huang and Riloff \cite{HuangRiloff-50}		& joint	& CRFs	& MUC-4	& cross-sentence	& general IE\\
		2012	& Sakaki et al. \cite{SakakiMatsuo-5}		& pipeline	& SVM	& Dataset \cite{SakakiMatsuo-5}		& document-level	& intelligent transportation\\
		2011	& Björne and Salakoski \cite{BjorneSalakoski-57}		& pipeline	& SVM	& BioNLP’11	& sentence-level	& biomedical\\
		2011	& Hong et al. \cite{HongZhang-51}		& pipeline	& SVM	& ACE2005	& sentence-level	& general IE\\
		2010	& Björne et al. \cite{BjorneGinter-56}		& pipeline	& CRF, SVM	& PubMed	& sentence-level	& biomedical\\
		2010	& Ananiadou et al. \cite{AnaniadouPyysalo-62}		& survey	& -	& -	& -	& biomedical\\
		2010	& Miwa et al. \cite{MiwaSaetre-55}		& pipeline	& SVMs	& BioNLP’09	& sentence-level	& biomedical\\
		2010	& Sætre et al. \cite{SaetreYoshida-54}		& pipeline		& AkaneRE & BC-IPT,  BioNLP'09	& sentence-level	& biomedical\\
		2010	& Li et al. \cite{LiLiu-61}		& pipeline	& CRF, AdaBoost, SVM	& i2b2  \tablefootnote{\url{https://www.i2b2.org/NLP/DataSets/Main.php}} 	& sentence-level	& biomedical\\
		2010	& Liao and Grishman \cite{LiaoGrishman-47} 		& pipeline	& ME	& ACE	& sentence-level	& general IE\\
		2009	& Patwardhan and Riloff \cite{PatwardhanRiloff-48}		& joint	& Naive Bayes, SVM	& MUC-4, ProMed	& document -level	& general IE\\
		2008	& Tanev et al. \cite{TanevPiskorski-4}		& pipeline	& Pattern Matching \& clustering	& Dataset \cite{TanevPiskorski-4}		& cross-document	& security monitoring\\
		2008	& Ji and Grishman \cite{JiGrishman-49}		& pipeline	& ME	& ACE2005, TDT	& cross-document	& general IE\\
		2006	& Naughton et al. \cite{NaughtonKushmerick-53}		& pipeline	& HAC 	& Dataset \cite{NaughtonKushmerick-53}		& cross-document	& general IE\\
		2006	& Ahn \cite{Ahn-52}		& pipeline	& TiMBL  \tablefootnote{\url{http://ilk.uvt.nl/timbl/}}, MegaM  \tablefootnote{\url{http://users.umiacs.umd.edu/~hal/megam/version0_3/}}	& ACE2005	& sentence-level	& general IE\\
		\bottomrule 
	\end{tabular}
	\label{tab2}
\end{table}

The features reported in previous machine learning based event extraction methods can be categorized into lexical and contextual features. Lexical features contain part-of-speech tags (POS), entity information, and morphology features (e.g., token, lemma, etc.) \cite{ChenXu-3}. Contextual features include local information (sentence level), global information (document level), and external dictionaries. These features are complementary, and there have been various research combining global evidence from related documents with local decisions \cite{LiaoGrishman-47, PatwardhanRiloff-48, JiGrishman-49}. For example, to overcome the shortage of analyzing sentences in isolation, Huang and Riloff \cite{HuangRiloff-50} present a bottom-up architecture to consider a view of the larger context. It is implemented by integrating sequential sentence classifiers that capture textual cohesion, including lexical associations and discourse relations across sentences. To resolve the ambiguities of sentence-level event extraction relying on local information, Liao and Grishman \cite{LiaoGrishman-47} use document-level statistical information to improve sentence-level event extraction to achieve document level within-event and cross-event consistency. Patwardhan and Riloff \cite{PatwardhanRiloff-48} combine phrasal and sentential evidence into a probabilistic framework to enhance accuracy. Hong et al. \cite{HongZhang-51} use blind cross-entity inference to improve sentence-level ACE event extraction by considering the consistency and distribution of entities and roles.

Considering the complexity of the event extraction task, the foremost researchers divide the task into four subtasks: event trigger identification, event type classification, argument detection, and role classification. There is much research to train the classifiers in a pipelined manner, with the advantage that the previous classifier can provide information to later classifiers \cite{Ahn-52, NaughtonKushmerick-53, JiGrishman-49, TanevPiskorski-4, LiaoGrishman-47, LiuHuang-37, SaetreYoshida-54, MiwaSaetre-55, BjorneGinter-56, HongZhang-51, BjorneSalakoski-57, SakakiMatsuo-5, PengMoh-13, HennSticha-16}. For example, Peng et al. \cite{PengMoh-13} propose an automatic pipeline to extract adverse drug events (ADE) by using Naïve Bayes and Support Vector Machine (SVM) to detect drug-related tweets and sentiment analysis before mapping the biomedical text into drug events. However, the shortcoming of pipelined training is also obvious: error propagation (cascading defects). To deal with this problem, researchers adopt a joint training manner that treats the event extraction task as a multi-classification problem \cite{PatwardhanRiloff-48, HuangRiloff-50, LiZhu-58, ChenNg-59, LuRoth-60}. For example, Chen and Ng \cite{ChenNg-59} employ joint learning for Chinese event extraction and investigate (1) various linguistic features that exploit results of zero pronoun resolution and noun phrase coreference resolution, and (2) features that exploit trigger probability and trigger type consistency.

From the technique view, the support vector machine (SVM), maximum entropy (ME), Naive Bayes (NB), conditional random field (CRF), integer logic programming (ILP), Hierarchical agglomerative clustering (HAC) are the most used machine learning algorithms. Lu and Roth \cite{LuRoth-60} present a semi-Markov CRF approach for automatic event extraction and further develop a novel learning approach called PM (structured preference modeling) that allows structured knowledge to be incorporated effectively in a declarative manner. Björne and Salakoski \cite{BjorneSalakoski-57} use SVMs to extract biomedical events (detailed descriptions of biomolecular interactions) from research articles in a pipelined manner.

From the application field view, these machine learning based event extraction models evolve in many areas, including general information extraction \cite{Ahn-52, NaughtonKushmerick-53, JiGrishman-49, LiaoGrishman-47, HongZhang-51, HennSticha-16, HuangRiloff-50, LiZhu-58, ChenNg-59, LuRoth-60}, biomedical \cite{LiLiu-61, SaetreYoshida-54, MiwaSaetre-55, BjorneGinter-56, BjorneSalakoski-57, PengMoh-13}, intelligent transportation \cite{SakakiMatsuo-5}, security monitoring \cite{TanevPiskorski-4}, etc. For example, Sakaki et al. \cite{SakakiMatsuo-5} develop a system that extracts real-time driving information using social media to offer important events to drivers, such as traffic jams and weather reports. It is beneficial for areas where Intelligent Transportation System (ITS) deployment is poor. In the security field, Tanev et al. \cite{TanevPiskorski-4} perform real-time news event extraction for global crisis monitoring. Many research efforts were centered around BioNLP event extraction shared tasks, e.g., extracting protein interactions from text \cite{SaetreYoshida-54}. Li et al. \cite{LiLiu-61} incorporate three supervised machine learning models: CRF, AdaBoost, and SVM, to automatically extract medication events from clinical text. Bj\"{o}rne et al. \cite{BjorneGinter-56} study the feasibility of performing event extraction at the PubMed scale. Miwa et al. \cite{MiwaSaetre-55} construct a model for extracting complex biomolecular events, e.g., binding and regulation, using rich features. Ananiadou et al. \cite{AnaniadouPyysalo-62} give a review of the current event extraction methods for systems biology.

There is much research involving in the specific domain or improving the extraction accuracy. Henn et al. \cite{HennSticha-16} perform case studies on how visualization techniques enhance automated event extraction. Naughton et al. \cite{NaughtonKushmerick-53} merge and extract events from heterogeneous news sources. There is also much research involving other language event extraction, for example, Chinese event extraction \cite{ZhaoQin-63, ChenJi-64, LiZhu-58}. Li et al. \cite{LiZhu-58} employ joint learning for Chinese event extraction and solve the high ratio of pseudo trigger mentions to true ones by using trigger filtering schemas. 

We end this section by summarizing the advantages and disadvantages of machine learning based event extraction by comparing it with pattern matching based methods. The benefits lie in two folds. Machine learning methods alleviate much effort to design delicate patterns and have better generalization and reusability. The disadvantages lie in three folds. First, supervised methods need more labeled data to train the model. Second, Feature engineering is a time-cost but critical step that affects extraction accuracy. Third, traditional machine learning methods have limitations in learning deep or complex nonlinear relations.

\subsection{Deep Learning based Methods}

Feature engineering is the main challenging issue of traditional event extraction methods. And traditional machine learning methods have limitations in learning deep or complex nonlinear relations. Deep learning based methods can alleviate these shortages due to their two distinguishing characters. First, embedded representation of input is suitable for big data. Second, specific deep architectures can better capture various more complex nonlinear features. This section first reviews the recent deep learning based event extraction literature, then summarizes it in Table \ref{tab3} from the year, model, paradigm, technique, datasets used, event-level performed, and application area. Then we focus on discussing the common characters of typical research from the view of feature, technique, and application field, without much effort in describing the details of specific methods. We finally summarize the advantages and disadvantages of deep learning based event extraction methods.

\begin{table}
	\caption{Summary of representative deep learning based event extraction methods.}
	\centering
	\scriptsize
	\begin{tabular} {llllllll} 
		\toprule 
		Year	& Model	& Paradigm	& Techniques	& Datasets 	& Event Level	& Application Field\\
		\midrule 
			2021	& TEXT2EVENT \cite{LuLin-65}	& joint	& Transformer, Encoder-Decoder	& ACE2005, ERE	& sentence-level	& general IE\\
			2021	& GATE \cite{AhmadPeng-66}	& joint	& GCN, Transformer	& ACE2005	& sentence-level	& cross-lingual\\
			2021	& Zhao et al. \cite{ZhaoZhang-67}	& joint	& GCN, hypergraph, HANN	& MLEE  \tablefootnote{\url{http://nactem.ac.uk/MLEE/\#availability}}, GE11  \tablefootnote{\url{https://www.kaggle.com/nishanthsalian/genia-biomedical-event-dataset}}	& document-level	& biomedical\\
			2021	& CACT \cite{LybargerOstendorf-68}	& joint	& Bert, Bi-LSTM	& CACT \cite{LybargerOstendorf-68}	& sentence-level	& biomedical\\
			2021	& Lybarger et al.\cite{LybargerOstendorf-69}	& joint	& Bert, Bi-LSTM, Attention	& SHAC\cite{LybargerOstendorf-69}, MIMIC, UW	& sentence-level	& social\\
			2021	& Min et al. \cite{MinRozonoyer-70}	& pipeline	& Bert, Pooling, 	& Min et al. \cite{MinRozonoyer-70}	& sentence-level	& social\\
			2021	& CasEE \cite{ShengGuo-17}	& joint	& Bert, Decoder	& FewFC	& sentence-level	& financial\\
			2021	& PROTEST-ER \cite{CaselliMutlu-71}	& joint	& Retraining BERT	& ProtestNews  \tablefootnote{\url{https://emw.ku.edu.tr/clef-protestnews-2019/}}	& sentence-level	& political\\
			2021	& EEOK \cite{MaoLi-72}	& pipeline	& DA-EvtLSTM, EvtLSTM	& EEOK 	& cross-sentence	& social\\
			2020	& Filtz et al. \cite{FiltzNavas-Loro-14}	& multimodel	& Bert, CRF, rule-based	& ECHR  \tablefootnote{\url{http://ilk.uvt.nl/timbl/}}	& sentence-level	& legal\\
			2020	& DEED \cite{HuangPeng-73}	& joint	& Deep Value Networks	& ACE2005	& document-level	& general IE\\
			2020	& RCEE \cite{LiuChen-74}	& MRC	& Bert	& ACE2005, FrameNet	& sentence-level	& general IE\\
			2020	& Wei et al. \cite{WeiJi-11}	& joint	& Bi-LSTM-CRF, CNN-RNN	& n2c2	& sentence-level	& biomedical\\
			2020	& GAIA \cite{LiZareian-75}	& pipeline	& ELMo, LSTM, CRF, CNN 	& SM-KBP2019  \tablefootnote{\url{https://tac.nist.gov/2019/SM-KBP/index.html}}	& sentence-level	& multimedia\\
			2019	& Doc2EDAG \cite{ZhengCao-6}	& joint	& Transformer, BI-LSTM-CRF	& ChFinAnn4  \tablefootnote{Crawling from \url{ http://www.cninfo.com.cn/new/index}}	& document-level	& financial\\
			2019	& PLMEE \cite{YangFeng-76}	& pipeline	& Bert	& ACE2005	& sentence-level	& general IE\\
			2018	& JMEE \cite{LiuLuo-77}	& joint	& GCN, Attention	& ACE2005	& sentence-level	& general IE\\
			2018	& TEES-CNN \cite{BjorneSalakoski-78}	& pipeline	& CNN	& BioNLP, GENIA	& sentence-level	& biomedical\\
			2016	& JRNN \cite{NguyenCho-79}	& joint	& RNN	& ACE2005	& sentence-level	& general IE\\
			2015	& DMCNN \cite{ChenXu-3}	& pipeline	& CNN	& ACE2005	& sentence-level	& general IE\\
		\bottomrule 
	\end{tabular}
	\label{tab3}
\end{table}

\begin{figure}[htbp]
	\centering
	\includegraphics[width= 6.5 in]{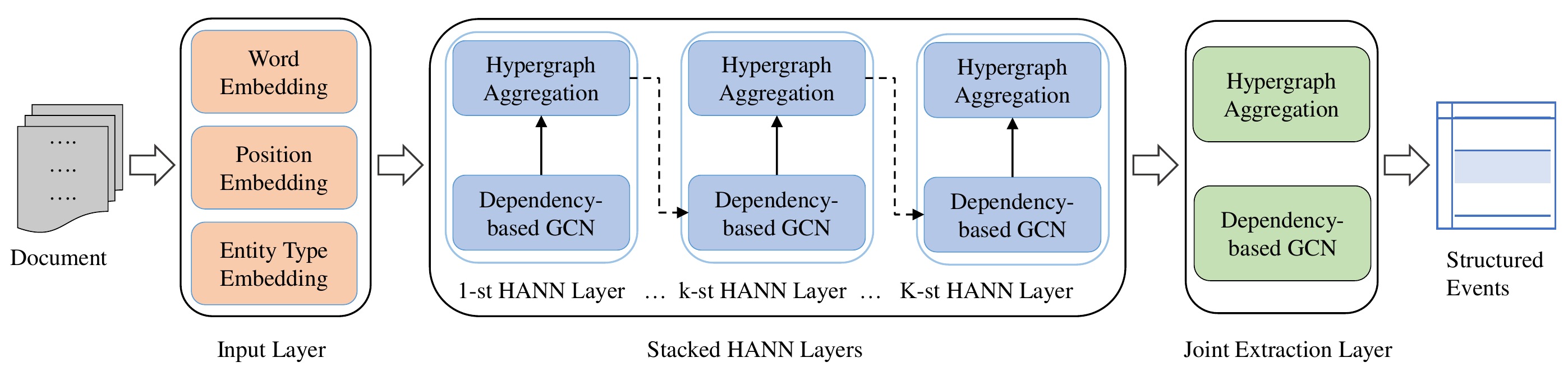}	
	\caption{Overview of the proposed framework of Zhao et al. \cite{ZhaoZhang-67}.}
	\label{fig6}	
\end{figure}

Deep learning methods can learn distributed representation of knowledge, e.g., semantic features, avoiding feature engineering. Word embedding, character embedding, position embedding, entity type embedding, POS tag embedding, entity type embedding, word distance, relative position, path embedding, etc., are the most used features \cite{ZhaoZhang-67, BjorneSalakoski-78, LiuLuo-77}. Except for the multi-channel distributed representation of the input, researchers have employed some techniques to capture the features contained in these representations. For example, to better capture the complex relationships among local and global contexts in biomedical documents, Zhao et al. \cite{ZhaoZhang-67} use a dependency-based GCN network to capture the local context and a hypergraph to model the global context. In addition, the fine-grained interaction between the local and global contexts is captured by a series of stacked Hypergraph Aggregation Neural Network (HANN) layers. The overview of the proposed framework is shown in Figure \ref{fig6}. 

Most recent event extraction studies are based on deep learning techniques, such as CNN \cite{ChenXu-3, BjorneSalakoski-78, LiZareian-75}, LSTM \cite{WeiJi-11, MaoLi-72}, Transformer \cite{ZhengCao-6, LuLin-65, AhmadPeng-66}, GCN \cite{AhmadPeng-66, ZhaoZhang-67, LiuLuo-77}, Bert \cite{LiuChen-74, ShengGuo-17, MinRozonoyer-70}, etc. There are also many hyhrid methods integrating the mentioned architectures to obtain super performance \cite{ZhengCao-6, WeiJi-11, LybargerOstendorf-69}. We group the mentioned research by used techniques and give short introductions of the typical works, respectively.
\paragraph{CNN based.}
CNN can capture local semantic features well in a sentence and overcome complex feature engineering compared with traditional machine learning methods \cite{ChenXu-3, BjorneSalakoski-78}. However, CNN may miss valuable facts when considering multiple-event sentences because it can not capture long-term information. Chen et al. \cite{ChenXu-3} use a dynamic multi-pooling convolutional neural network (DMCNN) to extract lexical-level and sentence-level features automatically. Björne and Salakoski \cite{BjorneSalakoski-78} use a CNN to capture a unified linear sentence representation, including semantic embeddings, position embeddings, and dependency path embeddings.

\paragraph{RNN \& LSTM based.}
RNN and LSTM architectures are good at capturing long-term and shot-term memory information, thus are suitable for sequence labeling and long dependency text. And event extraction can also be regarded as a sequence labeling task. For example, Nguyen et al. \cite{NguyenCho-79} use two bidirectional RNNs to learn a richer representation for the sentences. This representation is then utilized to predict event triggers and argument roles jointly. Wei et al. \cite{WeiJi-11} propose a Bi-LSTM-CRF-RNN-CNN approach to extract medications and associated adverse drug events (ADEs) from clinical documents. Specifically, in the named entity recognition phase, the BI-LSTM layers calculate scores of all possible labels for each token in a sequence. Then the CRF layer predicts a token's label using its neighbor's information. In the relation classification phase, all possible candidate relation pairs are generated by a structure that integrates CNN and RNN. To deal with the error propagation issue, Wei et al. \cite{WeiJi-11} propose a joint method for medication and adverse drug event extraction.

\paragraph{Attention \& Transformer based.}
Attention mechanisms allow the deep learning models to learn the most important information and ignore the noises by allocating different weights to different embeddings. According to the object the attention mechanisms work on, there are word-level, sentence-level, document-level, and channel-level attentions. The Transformer is a multi-head self-attention architecture in essence. Much attention-based or Transformer-based event extraction research has emerged. For example, Zheng et al. \cite{ZhengCao-6} propose an end-to-end model, Doc2EDAG, which can generate an entity-based directed acyclic graph to fulfill the document-level event extraction. The difference between Doc2EDAG and the classic method, Bi-LSTM-CRF, is that Doc2EDAG employs the Transformer instead of the original encoder, LSTM. The Transformer layers are used to encode a sequence of embeddings by the multiheaded self-attention mechanism to exchange contextual information among the token sequence. Lu et al. \cite{LuLin-65} also propose a sequence-to-structure generation paradigm that can directly extract events from the text in an end-to-end manner. Compared with \cite{ZhengCao-6}, a distinguishing difference is that \cite{LuLin-65} uses the event schemas as constraints to control the event records generation.

\paragraph{GCN based.}
Multiple events existing in the same sentence, arguments of one event across more than one sentence, or document-level event extraction are all facing one challenge: long-range dependencies. A common solution to leverage dependency structures is using universal dependency parses. Syntactic Graph Convolution Networks (GCNs) with nodes representing tokens and edges representing directed syntactic arcs are helping alleviate this challenge. To handle the difficulty of multiple events existing in the same sentence, Liu et al. \cite{LiuLuo-77} propose a novel Jointly Multiple Events Extraction (JMEE) framework to jointly extract multiple event triggers and arguments by introducing attention-based GCN to model the dependency graph information. Ahmad et al. \cite{AhmadPeng-66} use a Graph Attention Transformer Encoder (GATE) to learn the long-range dependencies and apply it in cross-lingual relation ad event extraction.

\paragraph{Bert based.}
Pretrained semantic representations, such as EMLo, Bert, have been widely used in multiple NLP tasks and have shown performance improvements in various NLP tasks. Bert is a bi-directional transformer architecture model, which has been trained on massive corpora and has learned fairly good semantic representations conditioned on token context and remains rich textual information \cite{ShengGuo-17}. Recently, much research has used Bert pre-trained representation as shared textual input features. For example, Liu et al. \cite{LiuChen-74} explicitly cast the event traction task as a machine reading comprehension problem and use question-answering techniques to perform event extraction. Min et al. \cite{MinRozonoyer-70} propose an event extraction framework, ExcavatorCovid, which extracts COVID- 19 related events and relations between them from news and scientific publications. These events are used to build a Temporal and Causal Analysis Graph, which will help the government sort out the information and adjust the related policies timely. The framework use Bert, Pooling, and linear layers for extracting temporal and causal relations.

\paragraph{Other new methods.}
Except for the mentioned deep learning based models, new paradigms of event extraction have emerged, such as question-answering based approaches \cite{LiuChen-74}. For example, Liu et al. \cite{LiuChen-74} explicitly cast the event traction as a machine reading comprehension problem and use question-answering techniques to perform event extraction. Many works are adopting strategies to improve extraction accuracy \cite{HuangPeng-73}. Many existing models seldomly consider the relationships between the event mentions and the event arguments in different sentences. To handle this challenge, Huang and Peng \cite{HuangPeng-73} propose a document-level event extraction framework, DEED, leveraging Deep Value Networks (DVN) to capture the cross-event dependencies and coreference resolution.

From an application perspective, these deep learning based event extraction models involve in many areas, including general information extraction \cite{ChenXu-3, NguyenCho-79, LiuLuo-77, YangFeng-76, LiuChen-74, HuangPeng-73, LuLin-65}, biomedical, \cite{BjorneSalakoski-78, WeiJi-11, LybargerOstendorf-68, ZhaoZhang-67}, financial \cite{ZhengCao-6, ShengGuo-17}, multimedia \cite{LiZareian-75}, legal \cite{FiltzNavas-Loro-14}, social \cite{MaoLi-72, MinRozonoyer-70,LybargerOstendorf-69}, political \cite{CaselliMutlu-71}, cross-lingual \cite{AhmadPeng-66}, etc.

We close this section by summarizing the advantages and disadvantages of deep learning based event extraction by comparing it with traditional methods. Deep learning, essentially, is an extension and development of machine learning. So it has the same pros and cons as machine learning. Here, we focus on summarizing the distinguishing strengths and weaknesses. The benefits lie in three folds. First, deep learning methods have more powerful nonlinear expression ability and can capture more complex relations between features, avoiding much feature engineering. Second, each deep learning method has its specialty and strong point in capturing syntactic and semantic features. For example, LSTM and Transformer architectures are all skilled in capturing long-range dependencies. Third, pre-trained models, especially Bert, can ford excellent context information and have been widely used as standard input features. The weaknesses of deep learning methods are as follow. First, due to the complex deep architectures, deep learning based models mainly rely on huge labeled corpora to train the model. Second, numerous parameter settings may affect the performance, such as the learning rate, training epochs, etc. However, many researchers have explored semi-supervised and unsupervised learning methods to alleviate the difficulty in obtaining labeled corpus.

\subsection{Semi-Supervised and Distant Supervision Methods}
Most event extraction systems are trained with supervised learning and rely on a collection of annotated data. Due to the domain-specificity of tasks, event extraction systems must be retrained with new massive annotated data for each domain \cite{HuangRiloff-80}. However, human-labeled training data is expensive to produce. Recently, some researchers have explored new methods, such as semi-supervised and distant supervision methods, to automatically produce more training data. 

\paragraph{Semi-Supervised methods.}
Semi-supervised learning (SSL) has attracted considerable attention to help achieve strong generalization by making use of both unlabeled data and labeled data \cite{LiuZhao-12, ZhouZhong-81, FergusonLockard-82, GuptaPawar-83, HuangJi-84, MansouriNaderan-Tahan-85, MiyazakiKomatsu-86, ZhouChen-87, ChenWang-88}. Much research has used various SSL methods to help generate data or augment data for event extractions: role-identifying nouns \cite{HuangRiloff-80}, linear discrimination analysis \cite{MansouriNaderan-Tahan-85}, Vector Quantized Variational Autoencoder \cite{HuangJi-84}, multi-modal Generative Adversarial Network \cite{ChenWang-88}, etc. 

\begin{figure}[htbp]
	\centering
	\includegraphics[width= 6.5 in]{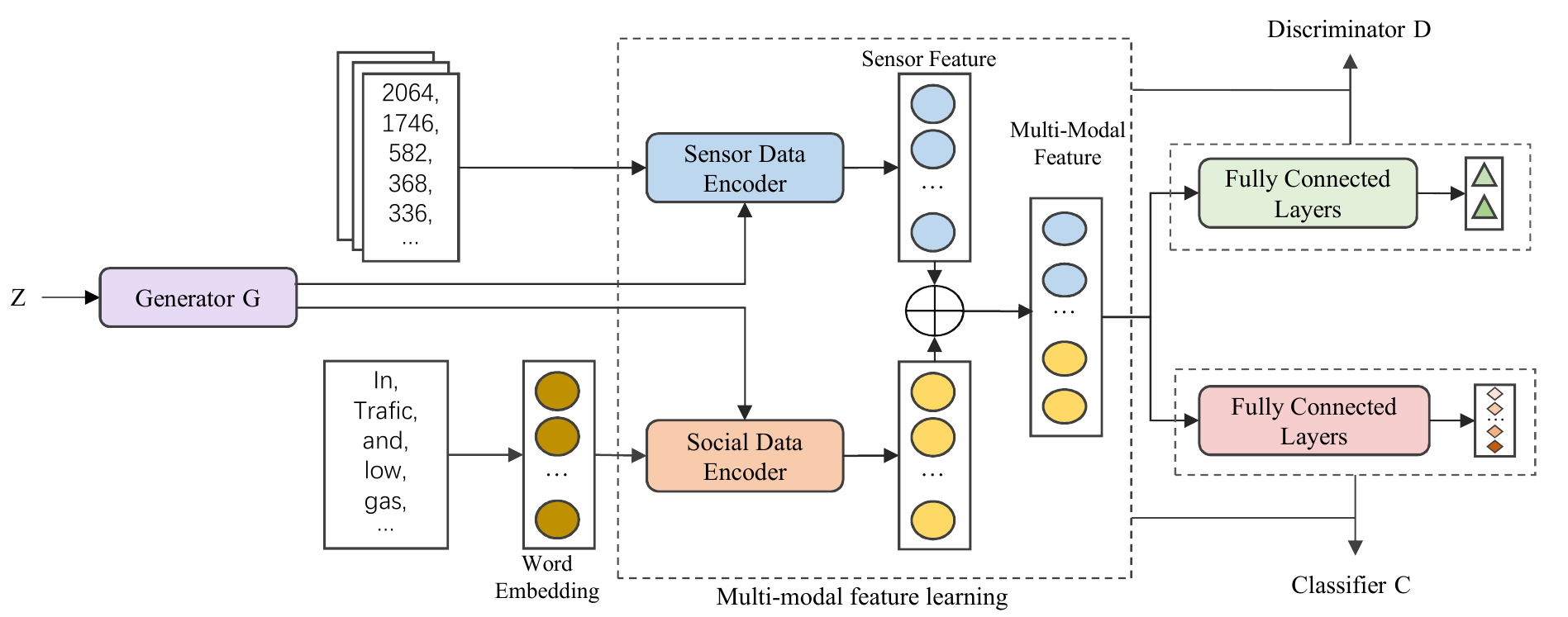}	
	\caption{Multi-modal feature learning from both sensor time series and text embeddings.}
	\label{fig7}	
\end{figure}

Huang and Riloff \cite{HuangRiloff-80} use role-identifying nouns to learn extraction patterns by a bootstrapping solution. Then the role-identifying nouns and patterns are used to create training data for event extraction classiﬁers. Mansouri et al. \cite{MansouriNaderan-Tahan-85} first use a convolutional neural network to extract explicit features from text and images, then use linear discrimination analysis (LDA) to predict the classes of unclassified data. Once the predicted accuracy is met, explicit features and predicted labels will be used to finally predict whether a piece of news is fake or real. The labeled and unlabeled instances are incorporated for training the semi-supervised learning model. Chen et al. \cite{ChenWang-88} extend the multi-modal Generative Adversarial Network (mmGAN) model to a semi-supervised architecture, which attempts to discriminate if the data is real or generated and categorize it into one of the two classes: traffic event or non-traffic event. As shown in Figure \ref{fig7}, the multi-modal feature learning architecture consists of three components: a Generator $G$, a Discriminator $D$, and a Classifier $C$. 

Different from the mentioned methods focusing on data generation and data augmentation, Zhou et al. \cite{ZhouChen-87} design a novel semi-supervised framework DualQA (dual question answering), to solve the event argument extraction in low-resource scenarios.

\paragraph{Distant Supervision Methods.}
Distant supervision is a successful paradigm that gathers training data for event extraction systems by automatically aligning vast databases of facts with the text \cite{ReschkeJankowiak-89, YangChen-90, ZuoChen-91, AlrashdiO'Keefe-92, BoudjellalZhang-93}. For example, Reschke et al. \cite{ReschkeJankowiak-89} present a new publicly available dataset and use the distant supervision approach to plane crash events. Yang et al. \cite{YangChen-90} first use Distance Supervision (DS) to automatically generate labeled data, then a sequence tagging model to extract document-level events from the financial announcements. The data generation contains two steps. First, the event trigger can be automatically marked by querying the pre-deﬁned dictionary (financial event knowledge base); thus, event mentions can be automatically identified, following the event trigger and the event arguments labeled. Second, once the event mention is identified, it is labeled as a positive example; then, the rest of the sentences in the announcement are marked as negative examples, which all constitute the document-level data. The deep event extraction architecture has a Bi-LSTM-CRF module for sentence-level and a CNN module for document-level event extraction. Zuo et al. \cite{ZuoChen-91} firstly design a Lexicon Enhanced Annotator (LexiAnno) to extract many causal event pairs based on linguistic knowledge and employ them to label sentences via distant supervision automatically. Experimental results show the proposed data augmentation framework outperforms other benchmark methods. To solve data lack and imbalance in coverage of crisis types, Alrashdi and O'Keefe \cite{AlrashdiO'Keefe-92} utilize distant supervision to automatically generate large-scale labeled tweet data for crisis response.

\subsection{Hybrid Methods}
Every single event extraction approach has its own merits and demerits. Combining different techniques can help integrate the advantages of multiple methods and significantly enhance the performance. There is an increasing number of researchers that employ multiple approaches, i.e., hybrid models. We review the existing literature and discuss it in two scenarios: single event extraction task and comprehensive system.

\subsubsection{Single Event Extraction Task}

\paragraph{Integrating Different Paradigm.}
As discussed above, we have divided the research into four paradigms: pattern matching methods, machine learning methods, deep learning methods, and data augmented methods. Many researchers have considered more than one paradigm to enhance the accuracy of event extraction. For example, Reschke et al. \cite{ReschkeJankowiak-89} extend the distant supervision approach to template-based event extraction and construct a new corpus, then use the linear-chain CRF model to test the performance on this dataset. Yang et al. \cite{YangChen-90} use the pattern-based methods to annotate the sentence-level and document-level corpus, then use the deep learning method to perform event extraction.

\paragraph{Integrating Different techniques.}
Because CRF and Bi-LSTM-CRF are widely used in different NER tasks, SVM and RNN-CNN are widely used in relation-classification tasks. And RNN is good at capturing global features, whereas CNN is good at capturing local features. Wei et al. \cite{WeiJi-11} propose a Bi-LSTM-CRF-RNN-CNN approach to extract medications and associated adverse drug events (ADEs) from clinical documents. Li et al. \cite{LiLiu-61} incorporate three supervised machine learning models: CRF, AdaBoost, and SVM automatically to extract medication events from clinical text.
GCN is good at modeling the long dependency parse, and Transformer is good at capturing the most important information. Ahmad et al. \cite{AhmadPeng-66} propose a deep model, integrating GCN and Transformer, to generate structured contextual representations based on the dependency parse results. 

The pre-trained models, such as Bert, can well represent the contextual semantic information and have been used as the standard input features. Then other deep learning architectures can be stacked based on this input layer, finetuned, and trained to execute related tasks. Lybarger et al. \cite{LybargerOstendorf-68} extract COVID-19 diagnoses and symptoms from clinical text. In this work, Bert, Bi-LSTM, Attention are used to generate pan representation. Specifically, firstly, Bert is used to map the input sentence into contextualized word embeddings. Then, these representations are feed to Bi-LSTM without finetuning the Bert. Lastly, each span is represented as the attention-weighted sum of the Bi-LSTM hidden states.

\subsubsection{Comprehensive System}
In recent years, event-related comprehensive systems have emerged. The remarkable character is that these systems extract multiple categorical information (e.g., entities, relations, and events), from multiple sources, multiple languages, and heterogeneous data modalities (speeches, texts, images, and videos). 

Li et al. \cite{LiZareian-75} present a comprehensive, open-source multimedia knowledge extraction system (GAIA) and create a coherent, structured knowledge base. This GAIA system enables the search of complex graph queries and retrieves multimedia evidence, including text, images, and videos. Specifically, the authors extract coarse-grained events and arguments using a Bi-LSTM-CRF model and a CNN-based model in the Text Knowledge Extraction (TKE) branch.

Wen et al. \cite{WenLin-18} also propose a comprehensive extraction system (RESIN) that can automatically construct temporal event graphs. RESIN extends from sentence-level event extraction to cross-document cross-lingual cross-media event extraction, coreference resolution, and temporal event tracking.

These event-related comprehensive systems have greatly enhanced the accuracy of information retrieval. The hybrid method integrates the advantages of multiple techniques, multiple sources, multiple languages, and heterogeneous data modalities, leading it to a mainstreaming paradigm in the future, especially in industrial applications.

\section{Open Domain Event Extraction}
The most distinguishing characteristic of open-domain event extraction is that it does not assume predefined event types and schemas. It usually focuses on detecting new or unexpected events \cite{VeysehNguyen-19, PengLi-10, ArakiMitamura-94}, event text generation \cite{FuBing-95, MartinAmmanabrolu-96}, specific field application ( e.g., energy prediction \cite{ChauEsteves-97}), and other general information extraction \cite{ShenZhang-98, NaikRose-23, LiuHuang-37, WangZhou-21, ZhouZhang-99, KunnemanVan-Den-Bosch-9, ZhouChen-8, RitterEtzioni-7, ArnulphyTannier-100}. In this section, we first review the recent open-domain event extraction literature and summarize it in Table \ref{tab4} from the view of year, model, paradigm, technique, datasets used, and subtasks. Then we categorize the literature into clustering-based, parsing-based, lexicon-based, semi-supervised \& distant supervision-based, Bayesian-based, Adversarial Domain Adaptation based, and open-domain event text generation from the view of technique. We focus on discussing the most distinctive characteristics of each typical research from the view of feature, technique, and application field, with little effort in describing specific methods' details. We finally summarize the advantages and disadvantages of open-domain event extraction methods. 

\begin{table}
	\caption{Summary of representative open-domain event extraction methods.}
	\centering
	\scriptsize
	\begin{tabular} {p{10pt} p{80pt} p{60pt} p{90pt} p{90pt} p{60pt} }
		\toprule 
		Year	& Model	& Paradigm	& Techniques	& Datasets 	& Tasks\\
		\midrule 		
		2021	& Veyseh et al. \cite{VeysehNguyen-19}	& semi-supervised	& Bert, BiLSTM	& LitBank, TimeBank, SW100	& Event Detection\\
		2021	& MONTEE \cite{de-VroeGuillou-22}
		& Lexicon \& parsing	& RotatingCCG parser, CoreNLP	& de Vroe et al. \cite{de-VroeGuillou-22}	& Event Fact Modality\\
		2021	& ETYPECLUS \cite{ShenZhang-98}
		& Parsing \& clustering	& Bert, EM, clustering algorithm	& ACE2005, ERE, Pandemic \tablefootnote{\url{https://github.com/mickeystroller/ETypeClus}}	& Event Type Induction\\
		2021	& Peng et al. \cite{PengLi-10}	& clustering	& GCN, DBSCAN	& Collected from Weibo \cite{PengLi-10} & Event Detection\\
		2020	& Chau et al. \cite{ChauEsteves-97}	& Lexicon \& parsing	& LSTM, Convolution	& Chau et al. \cite{ChauEsteves-97}	& Energy Prediction\\
		2020	& Naik and Rose \cite{NaikRose-23}	& Adversarial Domain Adaptation	& LSTM, Bert, POS	& LitBank, TimeBank	& General IE\\
		2020	& WAG \cite{FuBing-95}	& Encoder-Decoder	& RNN	& WikiEvent	& Story Generation\\
		2019	& ODEE \cite{LiuHuang-37}	& Bayesian	& ELMo, Bayesian, MLP	& GNBusiness	& General IE\\
		2019	& Dor et al. \cite{DorGera-101}	& weakly supervised	& Ruse based	& S\&P-wiki, Extended-wiki, SentiFM	& Fiancial\\
		2019	& AEM \cite{WangZhou-21}	& Bayesian	& Bayesian, GAN	& FSD, Twitter \cite{WangZhou-21}, Google  \tablefootnote{\url{http://data.gdeltproject.org/events/index.html}} 	& General IE\\
		2018	& Araki and Mitamura \cite{ArakiMitamura-94}	& unsupervised	& BiLSTM, WordNet	& SW100  \tablefootnote{\url{https://bitbucket.org/junaraki/coling2018-event}} 	& Event Detection\\
		2018	& Martin et al. \cite{MartinAmmanabrolu-96}	& Bayesian	& Bayesian, RNN	& movie plots \cite{BammanO-Connor-102}	& Story Generation\\
		2017	& DPEMM \cite{ZhouZhang-99}	& Bayesian	& Bayesian, LDA	& FSD, Twitter \cite{ZhouZhang-99}	& General IE\\
		2016	& Kunneman and Van D.B. \cite{KunnemanVan-Den-Bosch-9}	& clustering	& clustering	& Dutch tweets \cite{KunnemanVan-Den-Bosch-9}	& General IE\\
		2014	& LEM \cite{ZhouChen-8}	& unsupervised	& Bayesian	& FSD \cite{PetrovicOsborne-103}	& General IE\\
		2012	& TwiCal \cite{RitterEtzioni-7}	& supervised	& Linear Chain CRFs, Bayesian Inference	& Twitter \cite{RitterEtzioni-7}
		& General IE\\
		2012	& Arnulphy et al. \cite{ArnulphyTannier-100}
		& unsupervised	& rules, lexicon	& AFP \cite{ArnulphyTannier-100}, TimeBank	& General IE\\
		\bottomrule 
	\end{tabular}
	\label{tab4}
\end{table}

\paragraph{Clustering-based.}
Social events are the unique aggregation of various semantics, and related events or evolutions tend to be cohesive. Thus, density-based clustering algorithms can be used to detect new events and evolution discovery. For example, for each event group, an event schema can also be constructed with a slot-value schema through Event Schema Induction (ESI). Peng et al. \cite{PengLi-10} propose a streaming social event detection and evolution discovery framework. Specifically, first, an event-based heterogeneous information network (HIN) and a novel Pairwise Popularity Graph Convolutional Network (PP-GCN) are constructed. Then a parallel heterogeneous clustering algorithm (H-DBSCAN) is proposed for streaming event detection and evolution discovery.

\begin{figure}[htbp]
	\centering
	\includegraphics[width= 6.5 in]{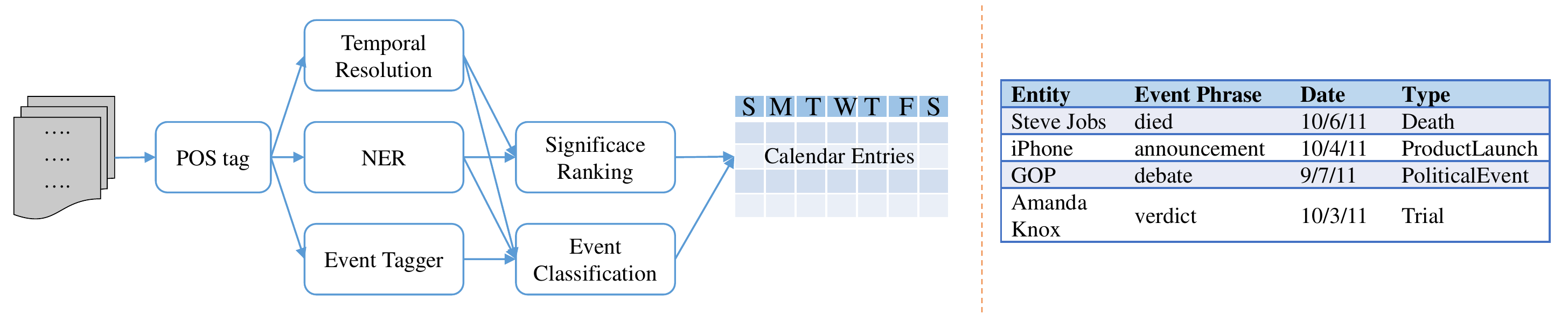}	
	\caption{Processing pipeline for extracting events from Twitter (left). The right side is examples of events extracted by TwiCal.}
	\label{fig8}	
\end{figure}

\paragraph{Parsing-based.}
Syntactic parsing results are widely used to enhance open-domain event extraction tasks. For example, the verb tag helps detect the event trigger, whereas the noun tag helps filter the event arguments. And the syntactic dependencies help catch the same event's roles and arguments, which appear across multiple sentences. Ritter et al. \cite{RitterEtzioni-7} present the ﬁrst open-domain event extraction and categorization system (TwiCal) for Twitter. As is shown in Figure \ref{fig8}, the processing pipeline contains POS tag, temporal resolution, NER, event tagger, significance ranking, and event classification components. Shen et al. \cite{ShenZhang-98} present an open-domain event type induction framework (ETYPECLUS). For this purpose, the framework first selects predicates and object heads, then disambiguates predicates, lastly induces the <hpredicate sense, object head> pairs by embedding and clustering algorithms. Chau et al. \cite{ChauEsteves-97} use syntactic parsing, WordNet, and a word sense disambiguation tool to extract events from news headlines. Then the events are used to feed to a deep neural network to predict the natural gas price.

\paragraph{Lexicon-based.}
Many researchers have contributed lexicons of words or phrases to assist the sequential event extraction tasks. For example, de Vroe et al. \cite{de-VroeGuillou-22} present an open-domain, lexicon-based event extraction system MONTEE that can distinguish different types of modality. It can tell a reported event had taken place, did not take place, or is uncertain. This result is valuable for avoiding extracting unreal events. Arnulphy et al. \cite{ArnulphyTannier-100} use patterns and shallow parsing to automatically build a lexicon for nouns event extraction.

\paragraph{Semi-supervised \& distant supervision based.}
Semi-supervised and distant supervision methods are able to generate high-quality training data. Veyseh et al. \cite{VeysehNguyen-19} explore a novel method for open-domain event detection by ﬁnetuning the pre-trained language model GPT-2 to automatically generate new training data. Particularly, a novel teacher-student architecture is adopted to keep the original and generated data consistency. Dor et al. \cite{DorGera-101} use rules to automatically extract weak labels for event mentions describing economic events. Araki and Mitamura \cite{FiltzNavas-Loro-14} use distant supervision to conduct open-domain event detection. The significant character is it can detect all kinds of events.

\paragraph{Bayesian-based.}
Most Bayesian-based open-domain event extraction models assume that a sentence or document is a joint distribution over event types, slots, entities, and contextual features. For example, Wang et al. \cite{WangZhou-21} propose an open event extraction model (AEM) based on Bayesian and Generative Adversarial Nets. Specifically, a Dirichlet prior and a generator are used to capture the patterns of latent events. In contrast, a discriminator is used to distinguish documents reconstructed from the latent events and the original input documents. Unlike other GAN-based text generation approaches that capture the generating text sequence, the generator in AEM learns the projection function between an event distribution and the event-related word distributions; thus, it captures the event-related patterns. Zhou et al. \cite{ZhouChen-8} propose a Bayesian model, called Latent Event Model (LEM), to extract a structured representation of events from social media. The most striking characteristic of LEM is that it is a fully unsupervised approach, and no annotated data is required. Reference \cite{LiuHuang-37} extracts event type, schema, and arguments using a neural latent variable network and Bayesian inference model (ODEE) and gets better results than other base models.

\paragraph{Adversarial Domain Adaptation.}
The adversarial domain adaptation (ADA) framework is initially proposed by Ganin and Lempitsky and has been widely used in multiple NLP tasks \cite{GaninLempitsky-104}. Naik and Rose \cite{NaikRose-23} leverage the adversarial domain adaptation (ADA) framework to identify event triggers. This framework treats the event trigger identification task as a token classification problem. A representation learner is trained to generate token-level representations, which are predictive for trigger identification but not for domain prediction, making it more domain-invariant. The obvious advantage is that there is no need to annotate the target domain data.

\paragraph{Open Domain Event Text Generation.}
Automated Story Generation (ASG) has been a research problem of interest and open-domain event extraction subtask. Fu et al. \cite{FuBing-95} perform an open-domain event text generation task with an entity chain as its skeleton. To build this dataset, a wiki augmented generator framework containing an encoder, a retriever, and a decoder is proposed. The encoder encodes the entity chain into hidden representations while the decoder decodes from these hidden representations and generates related stories. The retriever is responsible for collecting reliable information to enhance the readability of the generated text. Martin et al. \cite{MartinAmmanabrolu-96} model the automated story generation task as a sampling problem. It generates the following event by choosing the maximizing probability from the event distribution.

We close this section by discussing the advantages and disadvantages of the mentioned works compared with the closed-domain event extraction methods. Most open-domain event extraction works focus on detecting new events and extracting related information. This information is beneficial for scenarios that require comprehensive knowledge of broad-coverage, fine-grained, and dynamically evolving event categories, e.g., stock price prediction based on news. However, from the mentioned literature review, we can find that the existing methods are mainly based on syntactic parsing, clustering, Bayesian, lexicon, etc. The current methods' output is still not as perfect as closed-domain event extraction results in two aspects. First, since open-domain event extraction needs no predefined schemas, the extracted results are omnifarious, increasing the difficulty of utilization. Second, since open-domain event extraction has no predefined event types, some research uses the extracted event trigger to represent the event types. Although many researchers have tried to induce these event types by clustering or latent event type inference, the results are not always convenient or understandable. Due to the usefulness of dynamically evolving event categories, we believe that there will be more research exploring new paradigms and techniques in open-domain event extraction.

\section{Discussion}
In this section, we summarize and discuss current common research issues in event extraction. Despite the considerable progress in event extraction, there are still some challenges involved but not limited to the following aspects.

\paragraph{Datasets.}
Although there have been various annotated corpora and many researchers have explored some semi-supervised methods to automatically label data, the data size and categories still look embarrassed compared with the big data algorithm requirement. Another problem is category imbalance. For example, the existing corpus category mainly focuses on natural disasters, social relationships, biomedicine, etc. And the size of some categories are small. Even worse, there is no annotated corpus in some fields. More high-quality annotated data need more research, such as semi-supervised or distant supervision methods.

\paragraph{Document-level and corpus-level event extraction.}
Most existing event extraction methods mainly extract event arguments within the sentence scope \cite{ZhengCao-6, JiGrishman-49}. However, the extraction results are not ideal in the following two cases. First, event arguments of the same event always scatter across different sentences. Another, multiple sentences or documents characterize the same event. The former case leads the extraction results incomplete, while the latter case leads the extraction results redundant. Document-level and corpus-level event extraction tasks face the following challenges: long-term dependency and entity and event coreference. Researchers have started to settle this problem by various mechanisms, such as end-to-end structured prediction \cite{HuangPeng-73}, sequence-to-structure generation paradigm \cite{LuLin-65}, Open-schema event profiling \cite{YuanRen-105}, etc.

\paragraph{Cross-linguistic.}
Researchers have contributed relatively richer event extraction corpora in English, whereas fewer corpora in other languages. Recently cross-lingual transfer learning approaches have been used for event extraction \cite{SubburathinamLu-106, AhmadPeng-66}. For example, Subburathinam et al. \cite{SubburathinamLu-106} use a GCN-based network to train an event extraction model from source language annotations to the target language. However, GCN is not good at capturing long-range dependencies or not directly connected relations in the dependency tree. Ahmad et al. \cite{AhmadPeng-66} improve this work by using attention mechanisms to learn the dependencies between words with different syntactic distances. Cross-linguistic event extraction can save much effort in constructing corpora in other languages, and it is beneficial for low-source languages.

\paragraph{Event coreference.}
Usually, the same event frequently co-exists in multiple documents. For example, it is a common case that different news media report the same hot news. Even document-level event extraction may not alleviate the redundancy. Event coreference or event merge is crucial for information retrieval, especially in event-related comprehensive systems which involve multiple sources, multiple languages, and heterogeneous data modalities (speeches, texts, images, and videos). 

\paragraph{Open-domain event extraction needs new schemas and techniques.}
The current research primarily focuses on closed-domain event extraction due to its plentiful corpora, mature methods, and acknowledged evaluation mechanisms. Despite the importance of open-domain event extraction, it has not received sufficient attention compared with closed-domain event extraction research. We review the recent open-domain research and find that the performance has not reached the desired level. Several challenges still hinder its generalization and industrial applications. First, there are fewer large, high-quality, and acknowledged open-domain corpora. Second, a set of mature evaluation mechanisms to evaluate open-domain extraction results need to be proposed. Third, open-domain event extraction needs to develop new schemas and techniques to enhance the performance. We believe this is a promising research direction in future.

\section{Conclusion}
In this paper, we review and summarize the literature in event extraction from text. Overall, we focus on providing a comprehensive overview of event extraction tasks, ignoring the peculiarities of individual approaches. Specifically, we first introduce the related concepts of event extraction, such as EE catalog, task definition, corpora, evaluation metrics. Then we summarize the literature from the technique view. In both closed-domain and open-domain event extraction sections, we summarize the literature from the year, common framework, technique, corpus, application field, advantage, and disadvantage. Last, we summarize and discuss current common issues and related progress in closed-domain and open-domain event extraction. 

Although there are still many challenges, event extraction, especially open-domain event extraction, is attracting more and more attention due to its crucial role in information extraction. This research provides a way to quickly understand up-to-date event extraction tasks from a moderately difficult perspective.

\bibliographystyle{unsrt}  
\bibliography{eerefbib}

\end{document}